\documentclass[preprint,12pt]{elsarticle}

\usepackage{amssymb}
\usepackage{amsmath}
\usepackage{booktabs}

\usepackage{algorithm}
\usepackage{algorithmic}

\journal{}

\begin{document}

\begin{frontmatter}

\title{Intelligent Control of Spacecraft Reaction Wheel Attitude Using Deep Reinforcement Learning}

\author[label1]{Ghaith El-Dalahmeh\corref{cor1}}
\ead{geldalahmeh@swin.edu.au} 
\author[label1]{Mohammad Reza Jabbarpour}
\ead{rjabbarpoursattari@swin.edu.au} 
\author[label1]{Bao Quoc Vo}
\ead{bvo@swin.edu.au} 
\author[label2,label3]{Ryszard Kowalczyk}
\ead{Ryszard.Kowalczyk@unisa.edu.au} 

\affiliation[label1]{organization={Swinburne University of Technology},
             addressline={John St, Hawthorn},
             city={Melbourne},
             postcode={3122},
             state={VIC},
             country={Australia}}

\affiliation[label2]{organization={University of South Australia},
             city={Adelaide},
             postcode={5000},
             state={SA},
             country={Australia}}
\affiliation[label3]{organization={Systems Research Institute Polish Academy of Sciences},
             city={Warsaw},
             postcode={00-901}, country={Poland}
             }

\cortext[cor1]{Corresponding author: geldalahmeh@swin.edu.au}

\begin{abstract}
Reliable satellite attitude control is essential for the success of space missions, particularly as satellites increasingly operate autonomously in dynamic and uncertain environments. Reaction wheels (RWs) play a pivotal role in attitude control, and maintaining control resilience during RW faults is critical to preserving mission objectives and system stability. However, traditional Proportional-Derivative (PD) controllers and existing deep reinforcement learning (DRL) algorithms—such as TD3, PPO, and A2C—often fall short in providing the real-time adaptability and fault tolerance required for autonomous satellite operations. This study introduces a DRL-based control strategy designed to improve satellite resilience and adaptability under fault conditions. Specifically, the proposed method integrates Twin Delayed Deep Deterministic Policy Gradient (TD3) with Hindsight Experience Replay (HER) and Dimension-Wise Clipping (DWC) referred to as TD3-HD to enhance learning in sparse reward environments and maintain satellite stability during RW failures.

The proposed approach is benchmarked against PD control and leading DRL algorithms. Experimental results show that TD3-HD achieves significantly lower attitude error, improved angular velocity regulation, and enhanced stability under fault conditions. These findings underscore the proposed method’s potential as a powerful, fault-tolerant, on-board AI solution for autonomous satellite attitude control.

\end{abstract}


\end{frontmatter}
\section*{Nomenclature}

\begin{tabular}{l|l}
\textbf{ADCS} & Attitude Determination and Control System \\
\textbf{A2C} & Advantage Actor-Critic \\
\textbf{DWC} & Dimension-Wise Clipping \\
\textbf{DRL} & Deep Reinforcement Learning \\
\textbf{HER} & Hindsight Experience Replay \\
\textbf{LEO} & Low Earth Orbit \\
\textbf{MDP} & Markov Decision Process \\
\textbf{MRP} & Modified Rodrigues Parameters \\
\textbf{PD} & Proportional-Derivative \\
\textbf{PPO} & Proximal Policy Optimization \\
\textbf{RL} & Reinforcement Learning \\
\textbf{RW} & Reaction Wheel \\
\textbf{SAC} & Soft Actor-Critic \\
\textbf{TD3} & Twin Delayed Deep Deterministic Policy Gradient \\
\textbf{TD3-HD} & TD3 with Hindsight Experience Replay and DWC \\
\textbf{IS} & Importance Sampling \\
\end{tabular}

\section{Introduction}

The increasing demand for small satellites in low Earth orbit (LEO) for applications such as Earth observation, telecommunications, scientific research, and defense has underscored the importance of precise and resilient attitude control systems. In these missions, accurate orientation control is critical to ensure that the satellite's payload can perform optimally, collecting high-quality data, or maintaining stable communication links with ground stations \cite{mansell2020deep}. As small satellites operate in environments with frequent disturbances and limited resources, achieving high-performance attitude control has become a challenge. Traditional approaches, such as Proportional-Derivative (PD) controllers and their variants, have served as reliable solutions for standard attitude control tasks by correcting errors through simple proportional and derivative gains. However, these methods can be sensitive to changes in system dynamics, particularly in scenarios where hardware components, such as reaction wheels (RWs), fail or experience degradation \cite{mansell2021deep}. When such faults occur, traditional control strategies often lack the adaptability to compensate effectively, which can compromise overall mission success and, in severe cases, may render the satellite incapable of performing its intended functions.

RWs, key components of Attitude Determination and Control System (ADCS), provide fine-grained control over satellite orientation by adjusting the angular momentum. However, RWs are susceptible to wear, saturation, and degradation over time, which increases the likelihood of failures that can significantly compromise the stability of the ADCS \cite{mahfouz2021hybrid}. Such failures may result in reduced pointing accuracy, increased oscillations, and, in severe cases, complete loss of attitude control \cite{mansell2020deep}. Addressing these issues requires robust, fault-tolerant control mechanisms capable of adapting to RW performance degradation and potential failures. In scenarios involving unresponsive or degraded RWs, the limitations of conventional control strategies become evident as they struggle to adapt dynamically to compensate for these hardware issues. Recent advancements in artificial intelligence (AI) and, more specifically, reinforcement learning (RL), have introduced new possibilities for overcoming these challenges by offering adaptive, data-driven solutions that can respond to changing system states in real-time. RL-based methods are particularly suited for autonomous control applications, as they learn to make optimal decisions through trial and error, enabling them to handle unforeseen system changes, such as RW failures, in ways that traditional control systems cannot. This capability positions RL as a promising approach for enhancing spacecraft resilience and ensuring mission continuity despite component faults or operational uncertainties \cite{miralles2023critical}.

In light of these developments, this paper introduces a novel approach to attitude control for small satellites based on the Twin-Delayed Deep Deterministic Policy Gradient \cite{fujimoto2018addressing} with Hindsight Experience Replay \cite{andrychowicz2017hindsight} (TD3-HD) algorithm. The TD3-HD method leverages the strengths of twin-delayed gradients for stability in training and HER to efficiently handle sparse or challenging reward scenarios, making it well-suited for complex control tasks under dynamic fault conditions. Dimension-Wise Clipping (DWC) \cite{wu2022improved} is also integrated into the approach to further stabilize control by independently managing torque adjustments for each RW, preventing overcorrections or undercorrections that could destabilize the system in cases of partial actuator failures. The goal of this study is to assess the feasibility and benefits of applying TD3-HD in the context of ADCS, aiming to improve the resilience, adaptability, and control precision of the system under both normal operations and failure scenarios, such as when RWs become unresponsive. The key contributions of this paper are as follows.

\begin{itemize}
    \item Presenting an in-depth overview of existing control methods and RL-based approaches to assess their strengths and weaknesses in the context of spacecraft attitude control.
    \item Introducing TD3-HD as an advanced deep reinforcement learning (DRL) algorithm specifically tailored to address fault-tolerant attitude control in spacecraft, offering a robust alternative to traditional PD and DRL-based controllers.
    \item Providing extensive comparison between TD3-HD with other DRL algorithms, including Proximal Policy Optimization (PPO), Advantage Actor-Critic (A2C), and Twin-Delayed Deep Deterministic Policy Gradient (TD3), to assess their control performance under both nominal and failure scenarios.
\end{itemize}
This paper is organized as follows: Section 2 reviews related work on traditional and recent DRL methods for spacecraft attitude control, emphasizing fault tolerance and the limitations of DRL algorithms. Section 3 formulates the satellite attitude control problem in the context of an unresponsive RW, defining control objectives and fault-tolerant strategies. Section 4 introduces our proposed TD3-HD with DWC methodology, detailing the reward function design and objective function to manage RW faults. Section 5 describes the experimental setup using the Basilisk simulation framework to evaluate TD3-HD's fault-tolerance in unresponsive RW scenarios. Section 6 presents the results, comparing TD3-HD with PD, PPO, A2C, and standard TD3, demonstrating its superior stability and accuracy in maintaining control under fault conditions. Section 7 summarizes the findings, highlighting TD3-HD’s advantages in adaptability and precision over traditional and DRL-based approaches. Section 8 conclusion and future work. 

\section{Related Works}
\begin{figure}[H]
    \centering
    \includegraphics[height=7cm]{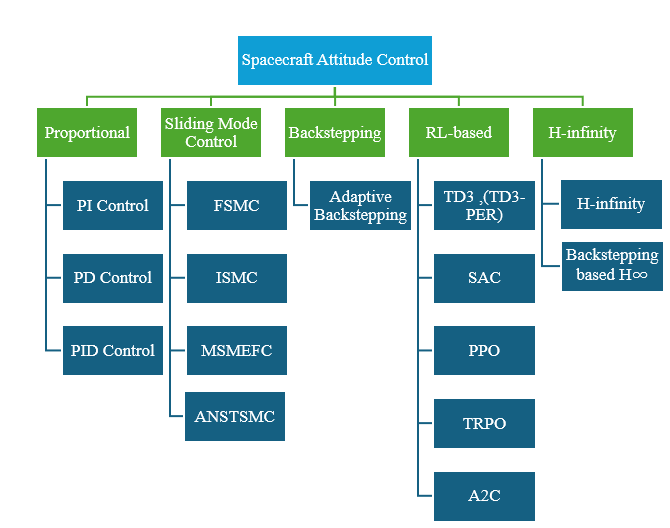}
    \caption{Overview of studied approaches}
    \label{fig:control}
\end{figure}

This section reviews the current state-of-the-art in spacecraft attitude control, focusing on traditional methods and recent advancements in RL-based control systems. An Overview of studied approaches is represented in Figure \ref{fig:control}. Relevant research on spacecraft control and autonomy is also discussed.

\subsection{Background}
 Various advanced control methods have been applied to spacecraft attitude control, alongside traditional methods like Proportional (P) controllers. These include Sliding Mode Control (SMC) for robustness, adaptive control for handling system variability, fuzzy logic for managing imprecise data, and neural networks for learning complex, and nonlinear dynamics. Each approach has shown significant promise, contributing to ongoing advancements in spacecraft resilience and autonomy.

\begin{itemize}
\item Proportional controllers:
Proportional-Derivative (PD) and Proportional-Integral-Derivative (PID) controllers have been widely used for spacecraft attitude control due to their simplicity and ability to achieve high pointing accuracy under stable conditions. However, as shown in Table~\ref{tab:proportional_comparison}, their performance is significantly constrained in the presence of disturbances, rapidly changing environmental conditions, or uncertainties in system parameters~\cite{henna2020towards}. These limitations arise from the use of fixed gains, which reduce their adaptability to nonlinearities and time-varying dynamics inherent in spacecraft operations~\cite{mehrjardi2014pd, mohan2023cubesat}.
\begin{table}[H]
    \centering
    \footnotesize 
    \renewcommand{\arraystretch}{1} 
    \setlength{\tabcolsep}{2pt} 
    \begin{tabular}{|p{2.5cm}|p{3.5cm}|p{3.5cm}|p{3.5cm}|p{0.8cm}|}
        \hline
        \textbf{Method} & \textbf{Advantages} & \textbf{Disadvantages} & \textbf{Comments} & \textbf{Ref.} \\
        \hline
        P Control & Simple, low cost, fast response & Cannot eliminate steady-state errors & Combine with PI or PD for better performance. & \cite{orozco2024attitude, wu2020application} \\
        \hline
        PI Control & Eliminates steady-state errors, higher precision & Slow response to rapid changes & Adaptive PI can reduce manual tuning. & \cite{wu2020application, mohan2023cubesat} \\
        \hline
        PD Control & Reduces overshoot, fast response & Cannot handle steady-state errors & Hybridize with PID or fuzzy logic. & \cite{mehrjardi2014pd} \\
        \hline
        PID Control & Simple structure, widely applied & Requires tuning, not robust to disturbances & Adaptive PID improves robustness. & \cite{wu2020application, parsai2019model} \\
        \hline
    \end{tabular}
    \caption{Comparison of Proportional controllers}
    \label{tab:proportional_comparison}
\end{table}
\item Sliding Mode Control:
 Sliding Mode Control (SMC) provides a robust solution to spacecraft attitude control, overcoming the limitations of traditional PID controllers by handling disturbances and uncertainties, such as actuator faults. As summarized in Table~\ref{tab:smc_comparison}, various SMC variants have been developed to mitigate chattering and enhance control performance. Adaptive SMC \cite{zhu2011adaptive} enhances system flexibility by estimating disturbances in real-time. However, despite its effectiveness, SMC suffers from chattering and high computational demands, limiting its applicability in systems with constrained resources.To mitigate these drawbacks, several enhancements have been introduced. Fuzzy Sliding Mode Control (FSMC), for instance, integrates fuzzy logic to reduce the chattering effect, as demonstrated in \cite{wang2023combined}. This integration improves control smoothness, although it increases computational complexity due to the need for precise tuning of fuzzy parameters. In a similar effort, Minimum Sliding Mode Error Feedback Control (MSMEFC) \cite{cao2014minimum} enhances control precision by minimizing sliding mode error. While this approach improves accuracy, it comes with the trade-off of increased computational load and heightened sensitivity to parameter tuning. In addition to these improvements, Adaptive Non-Singular Terminal Sliding Mode Control (ANSTSMC) \cite{modirrousta2017adaptive} tackles the chattering issue through adaptive tuning while avoiding singularities. Though ANSTSMC offers smoother control and finite-time convergence, its complexity restricts its real-time applicability in resource-constrained environments. Likewise, Adaptive Fuzzy Sliding Mode Control (AFSMC) \cite{xin2015adaptive} combines fuzzy logic with adaptive SMC, dynamically adjusting control laws to enhance tracking accuracy. Despite these improvements, the increased computational burden poses a challenge for systems with limited processing power. Furthermore, Integral Sliding Mode Control (ISMC) \cite{jia2020continuous} incorporates integral action into the SMC framework to improve steady-state accuracy in the presence of persistent disturbances. While this method demonstrates strong performance in managing uncertainties and actuator faults, its complexity and high processing requirements make real-time implementation difficult for systems with limited resources.
 In \cite{wang2019adaptive} a study that addressed the challenge of mitigating reaction wheel assembly (RWA) jitter in micro-satellites, which can significantly degrade attitude control accuracy due to their small moments of inertia. the authors proposed a combined approach using a Sliding Mode Controller (SMC) and an adaptive moment distribution algorithm to reduce jitter effects. The SMC ensures robust control under bounded disturbances, while the adaptive algorithm dynamically redistributes control torques based on wheel speeds, imposing limits to reduce vibration-induced disturbances. Simulation results demonstrate that this combination achieves high precision, reducing attitude errors to as low as 0.001°, a marked improvement over traditional PD controllers. However, the method relies on simplified jitter models, and requires careful tuning to balance speed limits and actuator saturation.

 In summary, although these advancements in SMC significantly improve robustness, control precision, and disturbance rejection capabilities, they introduce trade-offs in terms of complexity and computational demands. As a result, further optimization is necessary to make these methods more practical for real-time spacecraft applications, particularly in systems with constrained resources.
 \begin{table}[H]
    \centering
    \footnotesize
    \renewcommand{\arraystretch}{1.1}
    \setlength{\tabcolsep}{2pt}
    \begin{tabular}{|p{2.5cm}|p{3.5cm}|p{3.5cm}|p{3.5cm}|p{0.8cm}|}
        \hline
        \textbf{Method} & \textbf{Advantages} & \textbf{Disadvantages} & \textbf{Comments} & \textbf{Ref.} \\
        \hline
        SMC & Robust to disturbances, low computational cost & Causes chattering, actuator wear & FSMC, AFSMC, ISMC reduce chattering. & \cite{zhu2011adaptive} \\
        \hline
        FSMC & Handles uncertainties, reduces chattering & High computational cost & Hybrid methods optimize performance. & \cite{wang2023combined} \\
        \hline
        ISMC & Effective in magnetic attitude control & Cannot fully neutralize disturbances & Non-linear dynamics improve output control. & \cite{jia2020continuous} \\
        \hline
        ANSTSMC & Fast convergence, strong robustness & Computationally demanding & Suitable where precision outweighs cost. & \cite{modirrousta2017adaptive} \\
        \hline
    \end{tabular}
    \caption{Comparison of Sliding Mode Control Variants}
    \label{tab:smc_comparison}
\end{table}

\item Backstepping controller:
The researchers in \cite{freeman1993backstepping}, introduced Backstepping as a solution for nonlinear control systems, particularly effective for ensuring stability by dividing the system into smaller subsystems. This step-by-step approach improves trajectory tracking and system robustness. In space manipulators, Backstepping works alongside sliding mode control to maintain performance despite uncertainties and disturbances. As summarized in Table~\ref{tab:backstepping_comparison}, several variations of Backstepping have been proposed to address these challenges. For instance, \cite{babaei2023output} used Backstepping with the Modified Rodrigues Parameters (MRP) model and an extended state observer for satellite attitude control under disturbances. While effective, it increases computational complexity and introduces chattering. Similarly, \cite{yan2020backstepping} proposed an adaptive Backstepping fault-tolerant control (FTC) for handling actuator faults, though it faces issues with chattering and computational intensity. Lastly, \cite{boussadia2017adaptive} applied Backstepping to microsatellites with inertia uncertainties, showing good results but requiring fine-tuning and posing computational challenges for real-time use in resource-limited systems.

Despite its benefits, Backstepping can lead to high computational demands and the chattering effect, both of which hinder real-time implementation, especially in systems with limited processing power. Additionally, the need for precise tuning makes it challenging for adaptive and dynamic environments
\begin{table}[H]
    \centering
    \footnotesize
    \renewcommand{\arraystretch}{1.1}
    \setlength{\tabcolsep}{2pt}
    \begin{tabular}{|p{2.5cm}|p{3.5cm}|p{3.5cm}|p{3.5cm}|p{0.8cm}|}
        \hline
        \textbf{Method} & \textbf{Advantages} & \textbf{Disadvantages} & \textbf{Comments} & \textbf{Ref.} \\
        \hline
        Backstepping & Handles nonlinearities, Lyapunov stability & Computationally expensive & Adaptive techniques enhance flexibility. & \cite{babaei2023output} \\
        \hline
        Adaptive Backstepping & Adapts to uncertainties, high precision & High complexity, tuning required & ML/fuzzy systems improve real-time use. & \cite{boussadia2017adaptive} \\
        \hline
    \end{tabular}
    \caption{Comparison of Backstepping-Based Methods}
    \label{tab:backstepping_comparison}
\end{table}

\item \(H_{\infty}\) controller:
In \cite{zames1981feedback}, researchers developed the \(H_{\infty}\) control technique as a robust solution  to handle uncertainties, disturbances, and noise in control systems. It ensures system stability even when exact models of environmental conditions are unavailable, making it highly suitable for managing complex space systems. \(H_{\infty}\) control optimizes worst-case disturbance rejection, providing strong resilience in unpredictable environments. As outlined in Table~\ref{tab:robust_control_comparison}, various implementations of \(H_{\infty}\) control have been proposed to enhance spacecraft resilience. For instance, \cite{wang2023rigid} addressed attitude control in spacecraft with actuator misalignments and external disturbances using a nonlinear \(H_{\infty}\) controller, which ensures stability by solving the Hamilton–Jacobi–Isaacs (HJI) equation through the Chebyshev–Galerkin method. While effective, this approach's mathematical complexity can lead to conservative performance under nominal conditions.Similarly, \cite{esmaeilzadeh2023nonlinear} applied a backstepping-based nonlinear \(H_{\infty}\) controller to flexible spacecraft, achieving vibration suppression and finite-time stability during large-angle maneuvers. However, the method's intricate calculations and extensive parameter tuning present challenges for real-time implementation, especially in systems with limited resources.

Despite its robustness, \(H_{\infty}\) control's complexity leads to high computational demands and conservative performance in typical scenarios. These trade-offs highlight the need for further research to improve efficiency and adaptability for real-world satellite missions.

\begin{table}[H]
    \centering
    \footnotesize
    \renewcommand{\arraystretch}{1.1}
    \setlength{\tabcolsep}{2pt}
    \begin{tabular}{|p{2.5cm}|p{3.5cm}|p{3.5cm}|p{3.5cm}|p{0.8cm}|}
        \hline
        \textbf{Method} & \textbf{Advantages} & \textbf{Disadvantages} & \textbf{Comments} & \textbf{Ref.} \\
        \hline
        H-infinity & High robustness, performance bounds & Complex design, matrix equation solving & Efficient computational techniques help. & \cite{wang2023rigid} \\
        \hline
      
        backstepping-based  \(H_{\infty}\) & Energy-efficient, fault-resilient & Limited maneuver handling & Extend for orbital dynamics. & \cite{esmaeilzadeh2023nonlinear} \\
          \hline
    \end{tabular}
    \caption{Comparison of \(H_{\infty}\) Control Techniques}
    \label{tab:robust_control_comparison}
\end{table}

\end{itemize}
However, although these control techniques provide some level of resilience, they do so inefficiently due to their complexity and computational burdens. This inefficiency limits their practical use in real-time applications, particularly in small satellite systems. Further research is required to optimize these methods, making them both resilient and efficient for real-world, resource-constrained spacecraft missions. By addressing these limitations, future developments can aim to produce control solutions that are not only robust but also scalable and practical for real-time satellite operations. 
In recent years, RL has garnered substantial attention for its potential in spacecraft attitude control. Consequently, the following section provides a comprehensive overview of its application to spacecraft RW control, highlighting key methodologies, benefits, and challenges associated with this approach.

\subsection{RL in Spacecraft Attitude Control}
  
Artificial Intelligence (AI) has been acknowledged for its capability to address complex, real-world problems for many years \cite{davenport2018artificial}. While foundational deep learning techniques like backpropagation were developed in the 1990s \cite{werbos1990backpropagation}, their broader adoption was delayed until advancements in hardware and the availability of large datasets made them more feasible. Early attempts to apply AI to satellite control, such as those conducted at INPE \cite{carrara2000satellite}, demonstrated promise but were limited by the technological capabilities of the time. As AI and ML evolved, these technologies rekindled interest in intelligent control systems. RL, which shares a strong connection with optimal control theory \cite{bellman1954theory}, enables agents to learn behaviours through interaction with their environments, eliminating the necessity for precise mathematical models. As detailed in Table~\ref{tab:rl_methods_comparison}, various RL algorithms such as SAC, PPO, TD3, and A2C have been explored for their adaptability, robustness, and potential to handle nonlinearities and actuator faults.  Major advancements in RL, such as the introduction of Deep Q Networks \cite{mnih2015human}, significantly expanded its applicability to large state spaces by using neural networks to approximate the state-action value function, Q(s, a). Building on this foundation, the Deep Deterministic Policy Gradient (DDPG) algorithm \cite{lillicrap2015continuous} was introduced to handle continuous action spaces. Later advancements, including Twin-Delayed DDPG (TD3) \cite{fujimoto2018addressing} and Soft Actor-Critic (SAC) \cite{haarnoja2018soft}, addressed challenges like overestimation bias in DDPG, leading to enhanced training stability and efficiency. Furthermore, policy gradient methods such as Proximal Policy Optimization (PPO) \cite{schulman2017proximal} and Trust Region Policy Optimization (TRPO) \cite{schulman2015trust} became widely adopted due to their reliable and robust performance.
\cite{chai2018spacecraft} addresses the challenge of spacecraft attitude control under nonlinear dynamics, external disturbances, and model uncertainties, where traditional optimal control methods are limited by high computational demands and sensitivity to local minima. It proposes an analytical predictive controller based on Sequential Action Control (SAC), which calculates closed-form control actions within a receding horizon, eliminating the need for iterative optimization. The method is fast, robust to disturbances and model mismatches, easy to implement for jet thrusters or flywheels, and shows low sensitivity to initial conditions. However, the reliance on line search for action duration, which can be computationally inefficient. RL has seen extensive success in robotics, with notable applications in legged robots \cite{yang2020data} and robotic hand manipulation \cite{he2023robotic}. More recently, RL has been utilized in the aerospace domain. For instance,  \cite{hovell2020deep} combined RL for high-level guidance with a PD controller for spacecraft attitude control during docking maneuvers. The DRL approach may struggle with robustness against specific faults and often requires retuning of the PD controller to maintain stability. Sparse rewards can slow learning, requiring many interactions, and the method is computationally intensive, limiting real-time applicability. To address spacecraft attitude control during post-capture maneuvers with unknown dynamics, a Q-learning-based RL controller was proposed \cite{liu2022neural}, leveraging system input/output data to develop a model-free control strategy and eliminating the need for dynamic model identification. However, the Q-learning approach may struggle with robustness in uncertain environments due to the lack of a system dynamics model. Additionally, it is computationally intensive and slowed by sparse rewards, requiring numerous interactions to achieve reliable control in high-uncertainty settings. In \cite{meng4683974adaptive} authors proposed an adaptive fault-tolerant control (FTC) method for spacecraft using a Stackelberg game model integrated with Advantage Actor-Critic (A2C) RL. The approach addresses the challenge of maintaining spacecraft stability and efficiency under fault conditions by structuring FTC as a dynamic interaction between two players: a Fault-Tolerant Control (FTC) unit and a Fault Detection Observer (FDO). The A2C framework enables these units to adapt in real-time, improving resilience against faults. While promising, the solution requires high computational resources and may face scalability issues in multi-agent or complex environments, which the authors suggest as a focus for future research. Addressing the problem of control during debris removal missions is investigated in \cite{retagne2024adaptive}, where changes in mass distribution after debris capture render traditional PID controllers insufficient due to their reliance on fixed system parameters. To overcome this, the authors proposed a Deep Reinforcement Learning (DRL)-based adaptive control approach using algorithms like Soft Actor-Critic (SAC) and Proximal Policy Optimization (PPO) with a novel stacked observations technique. This method enables the system to infer dynamic properties, such as mass, indirectly from historical states and actions, improving adaptability across a wide mass range (10–1,000 kg). Simulations in the Basilisk framework demonstrate superior adaptability and robustness of the DRL-based system compared to PID controllers. However, still there is instability in extreme mass cases.
A recent study \cite{henna2024attitude} addresses a key challenge in RL-based spacecraft control by introducing a reward-shaping mechanism that guides agents using outputs similar to those of traditional PID controllers. This technique, termed expert-guided exploration (EGE), improves fault tolerance by efficiently directing the RL agent’s actions, particularly in environments with actuator or sensor malfunctions. By comparing RL and PID-like control outputs, EGE reduces the time needed to reach optimal policies, mitigating the risk of local minima and improving control robustness under uncertain conditions.While effective, the EGE approach's reliance on predefined similarity metrics may limit its full autonomy, as it necessitates expert input for optimal tuning and control vector adjustments. However \cite{henna2024attitude} also note that its computational complexity may impact real-time application feasibility for spacecraft with constrained onboard resources. A DRL-based approach for six degrees-of-freedom planetary landing, particularly for Mars exploration missions, was presented in \cite{gaudet2020deep}. The system employed PPO to map the estimated state of the lander directly to thrust commands, achieving both high accuracy and fuel-efficient trajectories. While the system showed robustness in simulations, The primary challenages of PPO include a high demand for samples, which makes training resource-intensive and time-consuming, and a sensitivity to hyperparameter tuning, impacting performance stability. While clipping aids stability, it can restrict exploration, sometimes trapping the agent in local optima. PPO also typically relies on multiple parallel environments for optimal results, adding to the computational load in complex applications. Recent advances in DRL have highlighted the potential of model-free approaches to enhance satellite attitude control, particularly in RW management. For instance, \cite{zhang2020model} presents a model-free attitude control solution for spacecraft using a PID-guided Twin-Delayed Deep Deterministic Policy Gradient (TD3) algorithm. This method seeks to address control challenges, such as external disturbances and control torque saturation, by employing RL techniques that enhance convergence speed and control stability. The incorporation of a PID guide for TD3 accelerates learning, potentially overcoming the slow training associated with TD3 in environments lacking prior knowledge. This study contributes to the growing application of RL in spacecraft control, demonstrating improved accuracy and adaptability under conditions of unknown dynamic parameters. However, as the authors noted that the deployment of the proposed solution in real-time space environments remains challenging due to the computational demands of pretraining and fine-tuning processes.
In \cite{sarmiento2023sample}, a Twin Delayed Deep Deterministic Policy Gradient with Prioritized Experience Replay (TD3-PER) was applied to the Diwata microsatellite, demonstrating significant improvements in sample efficiency and control stability. By leveraging prioritized experience replay, the study effectively reduced training time while maintaining precise attitude control, which is crucial for microsatellites operating under limited computational resources. However, TD3 with Prioritized Experience Replay (TD3-PER) is limited in handling partial actuator failures, such as an unresponsive RW, as it was designed for systems that face only environmental disturbances, not internal control failures, and lacks adaptive mechanisms for fault tolerance, leading to degraded control performance under fault conditions.  Motivated by the success of DRL in satellite attitude control, this study explores its application to address the critical challenge of unresponsive RW—a gap not fully covered in previous works. Managing unresponsive RWs is essential for maintaining satellite stability, and this research seeks to develop advanced control strategies to address such faults effectively. Standard TD3 has limitations in these scenarios, struggling with fewer successful outcomes, slow learning from failure cases, and limited exploration, which makes it less effective for controlling satellites with failed RWs. Building on these insights, the present study integrates HER with TD3, creating TD3-HD to enhance fault tolerance specifically in cases of unresponsive RWs. TD3-HD accelerates adaptation by enabling the agent to learn from unsuccessful attempts, addressing the shortcomings of standard TD3. Furthermore, DWC is incorporated to stabilize control by independently managing torque adjustments for each RW, preventing excessive or insufficient responses that could destabilize the system. This combined approach offers a novel solution to autonomously manage RW faults, providing a robust alternative to traditional methods such as PD control in the dynamic environment of low-Earth orbit.

\begin{table}[H]
    \centering
    \footnotesize 
    \renewcommand{\arraystretch}{1.1} 
    \setlength{\tabcolsep}{2pt} 
    \begin{tabular}{|p{2.5cm}|p{3.5cm}|p{3.5cm}|p{3.5cm}|p{0.8cm}|}
         \hline
        \textbf{Method} & \textbf{Advantages} & \textbf{Disadvantages} & \textbf{Comments} & \textbf{Ref.} \\
        \hline
        SAC & Adaptable to disturbances & High computation, needs stacked observations & Stacking improves performance. & \cite{chai2018spacecraft} \\
        \hline
        PPO & Stable learning, reliable control & Computationally demanding & Combine with other RL methods. & \cite{sarmiento2023sample} \\
        \hline
        TD3 & Stable, low energy use & Hyperparameter-sensitive & Delayed updates stabilize control. & \cite{zhang2020model} \\
        \hline
        DDPG & Learns complex policies efficiently & Prone to instability & Actor-Critic improves learning. & \cite{zhang2020model} \\
        \hline
        Q-learning & Model-free, adaptable to unknown dynamics & Computationally intensive, sparse rewards & Suitable for high-uncertainty environments. & \cite{liu2022neural} \\
        \hline
        A2C & Real-time adaptability, fault-tolerant & High computational resources, scalability issues & Promising for multi-agent systems. & \cite{meng4683974adaptive} \\
        \hline
        PPO stacked & Best overall performance & Adapts dynamically to mass variations & Slightly better at lower masses  More robust than PID but less consistent & \cite{retagne2024adaptive} \\
        \hline
        SAC stacked & More generalized approach & Adapts dynamically to mass variations & More consistent at higher masses  More robust than PID but never fully settles & \cite{retagne2024adaptive} \\
        \hline
        EGE & Guided exploration, fault-tolerant & Reliance on expert knowledge, computational overhead & Balances guidance and exploration. & \cite{henna2024attitude} \\
        \hline
        TD3-PER & High sample efficiency, stable control & Reward function tuning challenges & Optimizes resource-constrained systems. & \cite{sarmiento2023sample} \\
        \hline
         PPO & High accuracy, fuel-efficient trajectories & High sample demand, hyperparameter sensitivity & Robust for planetary landings. & \cite{gaudet2020deep} \\
        \hline
       
    \end{tabular}
    \caption{Comparison of RL Methods }
    \label{tab:rl_methods_comparison}
\end{table}

\section{Problem Formulation}
The spacecraft’s attitude is defined by how its body-fixed coordinate system aligns with the orbital coordinate system. The attitude describes the orientation of its body-fixed frame relative to the orbital frame. There are five common ways to describe attitude: Euler angles, Rotation Matrices, Quaternions, Rodrigues Parameters (RP), and Modified Rodrigues Parameters (MRP). These methods can be converted into one another, with their details available in references \cite{chaturvedi2011rigid}, \cite{bandyopadhyay2015attitude}. In this study, the body's orientation relative to the inertial frame is expressed using MRPs because they smoothly represent eigenaxis rotations up to 360 degrees without ambiguities or discontinuities \cite{calaon2022constrained} 

\subsection{Spacecraft Attitude Kinematics and Dynamics}

In the context of spacecraft kinematics and dynamics, satellite kinematics is the study of how a satellite’s orientation changes over time, focusing on its rotation without considering forces \cite{markley2014attitude}. The kinematic equation in terms of MRPs is given by \cite{su2011globally}:

\[
\dot{\rho} = G_\rho \omega \tag{1}
\]

where \( \omega \in \mathbb{R}^{3 \times 1} \) is the angular velocity vector relative to the inertial frame, and \( G_\rho \) represents a kinematic transformation matrix that relates the time derivative of the MRPs, \( \dot{\rho} \), to the spacecraft's angular velocity vector \( \omega \) and calculated as follows:

\[
G_\rho = \frac{1}{2} \left( I - S(\rho) + \rho \rho^T - \frac{1}{2} \rho^T \rho I \right), \tag{2}
\]

where \( I \) is the identity matrix of appropriate dimensions, \( S(\cdot) \) represents a skew-symmetric matrix. The MRPs are defined as \( \rho_i = q_i / (1 + q_0) \), where \( \bar{q} = [q_0, q^T]^T \in \mathbb{R} \times \mathbb{R}^3 \) represents the spacecraft quaternions, and \( q = [q_1, q_2, q_3]^T \). Thus,  MRPs are \( \rho = [\rho_1, \rho_2, \rho_3]^T \). 

While satellite dynamics focuses on a satellite's motion, including orientation changes (kinematics) and the effects of forces to ensure stability and control in space \cite{markley2014attitude}, the rotational motion of a rigid spacecraft with fully functioning actuators is governed by the following dynamics equation \cite{su2011globally,hughes2012spacecraft}:
\[
J \dot{\omega}_t - S(\omega_t) J \omega_t = u_t \tag{3}
\]
In this equation, \( J \in \mathbb{R}^{3 \times 3} \) is the spacecraft’s moment of inertia matrix, which is symmetric and positive-definite, reflecting how the mass is distributed about its three principal axes. The term \( \omega_t \in \mathbb{R}^3 \) represents the spacecraft’s angular velocity vector at time \( t \), and \( \dot{\omega}_t \) is its time derivative—i.e., the angular acceleration.

The term \( S(\omega_t) J \omega_t \) accounts for gyroscopic torques (Coriolis effects), where \( S(\omega_t) \) is the skew-symmetric matrix that performs the cross product operation with \( \omega_t \). This captures the nonlinear coupling between rotational axes during angular motion.

The right-hand side, \( u_t = [u_1,\ u_2,\ u_3,\ u_4] \in \mathbb{R}^4 \), represents the total control torque generated by four reaction wheels (RWs). Each RW contributes torque about a specific axis by changing its spin rate. The wheels operate within a rotational speed range of $[-1500, 1500]$ revolutions per minute and have a moment of inertia of $4.67 \times 10^{-4} \, \text{kg·m}^2$, enabling fine-grained control of the spacecraft’s angular momentum.

The control system aims to align the spacecraft’s current orientation represented using Modified Rodrigues Parameters (MRP)—with a desired target orientation. The difference between the current and target orientation is denoted as the attitude error \( \text{MRP}_{\text{error}} \), which is favored for its compactness and ability to avoid singularities.

To formally model and solve the spacecraft attitude control problem, a Markov Decision Process (MDP) formulation is introduced in the next section.

\subsection{Markov Decision Process Formulation }{
Spacecraft attitude control is framed as a Markov Decision Process (MDP). This formulation enables the application of Deep Reinforcement Learning (DRL) to develop optimal control strategies for maintaining the spacecraft’s orientation. Within this context, the spacecraft is modeled as an intelligent agent that interacts with its environment, adapting its actions to ensure stability and control even under challenging conditions, such as an RW failure.

The MDP is defined as:

\[
MDP = (S, A, T, P, r) \tag{4}
\]

A Markov Decision Process (MDP) is defined as a tuple \((S, A, T, P, r)\), where: The state space \(S\) represents all potential configurations of the system.  
The action space \(A\) encompasses all possible decisions the agent can make.  
The set \(T\), which is a subset of \(\mathbb{N}\), represents a sequence of time steps during which the agent interacts with the environment.  
The transition function \(P(s'|s, a)\) defines the probability of transitioning to a state \(s'\) given the current state \(s\) and action \(a\).  
The reward function \(r(s, a)\) specifies the immediate reward obtained by the agent when it takes action \(a\) in state \(s\).

The \textbf{state space} at time \( t \), denoted as \( s_t \), includes key parameters representing the spacecraft’s attitude error and angular velocity:
\[
s_t = \{\mathbf{MRP}_{\text{error}}, \mathbf{\omega}\} \tag{5}
\]
where:
\begin{itemize}
    \item \( \mathbf{MRP}_{\text{error}} \) represents the Modified Rodrigues Parameters that quantify the orientation error between the current and desired orientations. This choice enables the state vector to describe the attitude error relative to the target orientation in a simple, compact form, avoiding the complexity and potential singularities associated with other representations, such as Euler angles. By using MRPs, we can maintain a consistent, relative frame for error calculation, streamlining real-world implementation by removing any reliance on an arbitrary inertial reference frame \cite{elkins2020adaptive}.
    \item \( \mathbf{\omega}\) is the angular velocity of the spacecraft relative to an inertial frame, a parameter crucial for tracking the spacecraft’s rotational dynamics. Specifically, 
\( \mathbf{\omega} \)
  provides information on the instantaneous rotation rate of the spacecraft's body with respect to a fixed external reference. This velocity measurement allows the control system to make corrections that stabilize and adjust the spacecraft’s attitude as it seeks the target orientation. Including 
\( \mathbf{\omega} \)
  in the state vector is essential for capturing the spacecraft’s current motion state, enabling the agent to understand both the direction and magnitude of rotations. This, in turn, supports effective control strategies for damping oscillations and precisely orienting the spacecraft, particularly under conditions where reaction wheel limitations may impact stability
\end{itemize}

The \textbf{action space} \( a_t \) includes the control commands applied to the RWs:
\[
a_t = \{\tau_1, \tau_2, \tau_3, \tau_4\} \tag{6}
\]
where \( \tau_i \) is the torque applied to the  RW. In case of a fault where a RW becomes unresponsive, the control system must redistribute torque across the remaining functional wheels and activate a backup RW to maintain control and stability.
}
The goal of the control policy \( \pi(s_t) \) is to minimize attitude error and regulate angular velocity. To achieve this, we employ the TD3 (Twin Delayed Deep Deterministic Policy Gradient)  with HER and DWC, which stabilizes learning by clipping control inputs, especially in fault conditions that could destabilize the spacecraft.

\begin{figure}[H]  
    \centering
    \includegraphics[width=0.7\textwidth]{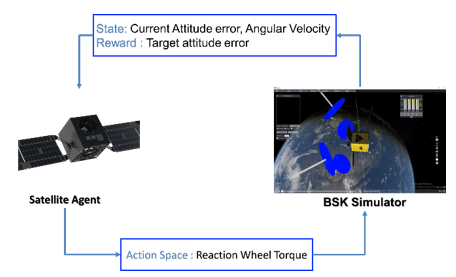}  
    \caption{Interaction of Agent with Environment via Basilisk}
    \label{fig:Simulation}
\end{figure}

To train the DRL agent and simulate the spacecraft's environment, we use the \textit{Basilisk} simulation framework. Basilisk provides a high-fidelity environment to simulate spacecraft dynamics and control subsystems, allowing for robust training of the DRL agent under dynamic and fault-prone conditions \cite{kenneally2020basilisk}. By simulating real-time spacecraft operations, including RW faults, As illustrated in Figure \ref{fig:Simulation}, Basilisk enables the agent to interact with the spacecraft’s control system, adapt its actions, and maintain stability. Using Basilisk as the simulation environment, this approach allows the spacecraft to autonomously manage faults, adapt control actions, and maintain attitude control in unpredictable space environments.

\section{Proposed Approach:}

In this section, we propose a \textbf{TD3-HD} approach to address the problem of spacecraft attitude control under unresponsive RW. When an RW becomes unresponsive, the proposed algorithm dynamically redistributes the torque to functional RWs and enables backup RW to compensate. The algorithm incorporates \textbf{Dimension-Wise Clipping (DWC)} to ensure stable action updates and \textbf{Hindsight Experience Replay (HER)} to improve learning efficiency in sparse reward environments.

\subsection{Twin-delayed deep deterministic policy gradient (TD3):}
TD3 is an off-policy, actor-critic, policy gradient algorithm that builds on Deep Deterministic Policy Gradient (DDPG) \cite{lillicrap2015continuous}, a popular method for continuous control tasks. However, DDPG often experiences instability and is highly sensitive to hyperparameter settings, primarily due to overestimated Q-values in the critic network. This overestimation can accumulate over time, leading the agent to converge to suboptimal solutions. TD3 addresses these challenges by introducing three major enhancements: (1) utilizing two critic networks, (2) delaying actor network updates, and (3) applying action noise regularization \cite{fujimoto2018addressing}.

\subsection{Hindsight Experience Replay (HER):}

\textbf{HER} effectively tackles the well-known challenge of sparse rewards in reinforcement learning by transforming unsuccessful episodes into valuable learning experiences \cite{fang2018dher}. In spacecraft attitude control tasks—especially under actuator faults such as unresponsive reaction wheels—the agent may receive minimal feedback if the reward is only granted upon full stabilization. This severely limits the agent's ability to learn effective control strategies early in training. HER overcomes this by retrospectively redefining the goals within each trajectory to match outcomes the agent actually achieved (e.g., intermediate attitude states), thereby converting failure into simulated success. This significantly increase the reward space, accelerating convergence and improving the agent's adaptability to fault scenarios.

Compared to other sparse reward mitigation techniques, HER stands out for its simplicity, generality, and effectiveness in goal-conditioned environments \cite{liu2021deep}. Alternatives like intrinsic motivation methods such as curiosity-driven exploration can promote broader state-space exploration but often lack task specific focus \cite{creus2024intrinsic}. Curriculum learning, where agents progress from easy to hard tasks, also shows promise but requires careful manual tuning and domain expertise to define appropriate training progressions \cite{narvekar2020curriculum}. In contrast, HER provides a plug-and-play solution that dynamically leverages the agent’s own experience to improve learning efficiency without requiring manual reward engineering or additional exploratory incentives. This makes HER particularly well suited for space applications, where stability and robustness under fault conditions are critical.

\subsection{Dimension-Wise Clipping (DWC):}
To address the problem of maintaining stability during updates in complex environments, the algorithm restricts updates within a defined range to prevent excessively large changes that could destabilize the system. However, this restriction can lead to zero-gradient problem \cite{zhang4663051gradient}. This issue becomes particularly problematic in high-dimensional action spaces, such as those required to control multiple reaction wheels, as it hinders effective learning \cite{zhang2022addressing}. To overcome this limitation, TD3-HD integrates Dimension-Wise Clipping (DWC), a technique introduced by \cite{han2019dimension}. 

In DWC, the policy gradient updates are clipped independently for each dimension of the action space. Specifically:

\begin{itemize}
    \item \textbf{Parameters clipped:} DWC applies independent clipping to each dimension of the policy gradient $\nabla_{\theta}J(\theta)$, where each dimension corresponds to an individual reaction wheel's torque adjustment.
    
    \item \textbf{Clipping mechanism:} For each dimension $i$, the gradient component $\nabla_{\theta}J(\theta)_i$ is clipped to a range $[-c_i, c_i]$, where $c_i$ is the clipping threshold for that dimension. This means each component is restricted to its own specific range, rather than applying a uniform clipping to the entire gradient vector.
    
    \item \textbf{Threshold determination:} The clipping thresholds $c_i$ are determined based on:
    \begin{itemize}
        \item Adaptive thresholds calibrated to the historical variance of gradients in each dimension
        \item Physical constraints of each reaction wheel, including maximum torque capacity
        \item Expected operational ranges for different control scenarios, including fault conditions
    \end{itemize}
\end{itemize}

DWC independently clips the torque adjustments for each RW, thereby preventing destabilizing updates without causing zero-gradient issues in unaffected dimensions. This preservation of learning signals in stable dimensions allows training to proceed efficiently even when some reaction wheels require significant adjustment while others need only minor corrections. This method significantly enhances the sample efficiency of TD3-HD by maintaining stability and responsiveness, even in the presence of faults. As a result, TD3-HD becomes highly effective for robust satellite attitude control, particularly in scenarios involving reaction wheel failures.
\subsection{Torque Redistribution}

When an RW becomes unresponsive, TD3-HD clips the corresponding action dimension, preventing further updates to that RW. The remaining active RWs receive updated actions based on the redistributed torque. The backup RW is activated only when necessary to ensure continuous control of the spacecraft’s attitude.

The total control torque \(T\) applied to the satellite is calculated as:
\[
T = \sum_{i=0}^{3} \lambda_i RW_i \tag{7}
\]
where \(RW_i\) is the torque applied by each RW, and \(\lambda_i\) is the policy parameter for each RW.

The parameter \(\lambda_i\) acts as a weighting factor that governs the contribution of each RW to the net control torque. These weights are dynamically adapted only for the currently active wheels. In nominal conditions, all active \(\lambda_i\) values are equal, assuming symmetric contribution. If an RW becomes faulty, its corresponding \(\lambda_i\) is set to zero, effectively excluding it from the control torque calculation. The remaining \(\lambda_i\) values are renormalized over the set of operational RWs to preserve control authority. When the backup RW is activated, it is assigned a non-zero \(\lambda_i\) and included in the redistribution scheme, ensuring seamless fault recovery and attitude control.

\subsection{Reward Function}

A reward function serves as a fundamental tool in reinforcement learning, guiding an agent's behavior toward achieving specific objectives \cite{arulkumaran2017deep}. It provides feedback by assigning a numerical value, or "reward," to each action or state, reflecting how well the agent's behavior aligns with the desired goals. The reward function plays a critical role in defining the learning objective and encouraging the agent to make decisions that optimize long-term outcomes. It effectively balances competing priorities, driving the agent toward desirable behavior while penalizing undesirable actions \cite{arulkumaran2017deep}. Our reward function is specifically designed to minimize attitude error while penalizing large angular velocities, comprising three components:

\begin{itemize}
    \item \textbf{Attitude Error Reduction Reward:}

    The reward function for attitude error reduction is defined as:
    \[
    r_1 = e_{\text{previous}} - e_{\text{current}} \tag{8}
    \]
    where:
    \begin{itemize}
        \item \( e_{\text{previous}} \) represents the attitude error at the previous time step. This reflects the difference between the spacecraft's desired and actual orientations at that earlier moment. It serves as a reference to determine if the attitude error is decreasing over time.
        \item \( e_{\text{current}} \) represents the attitude error at the current time step, calculated similarly to \( e_{\text{previous}} \). It reflects the current difference between the desired and actual orientations.
    \end{itemize}

    \item \textbf{Penalty for High Angular Velocities:}
    \[
    r_2 = 
    \begin{cases} 
    -10, & \text{if } |\omega| > 1 \\
    0, & \text{otherwise}
    \end{cases} \tag{9}
    \]
    where \( \omega \) is the angular velocity. High angular velocities are penalized to prevent instability and ensure safe control actions.

    \item \textbf{Accuracy Incentive:}
    \[
    r_3 = 
    \begin{cases} 
    0.01, & \text{if } e_{\text{current}} < 0.25 \\
    -0.01, & \text{otherwise}
    \end{cases} \tag{10}
    \]
    This component encourages precise attitude control by rewarding low attitude errors. The threshold value of 0.25 degrees was selected based on industry standards for high-precision satellite pointing requirements \cite{sawada2001high}. This value represents a balance between achievable physical performance given the hardware constraints of typical reaction wheel systems and mission requirements for Earth observation and scientific instruments. Experimental validation confirmed this threshold as optimal - lower values (e.g., 0.1 degrees) led to excessive control effort and oscillatory behavior without meaningful improvement in steady-state accuracy, while higher thresholds (e.g., 0.5 degrees) resulted in insufficient pointing precision for typical mission requirements. Additionally, the 0.25-degree threshold aligns with the capabilities of modern star trackers and inertial measurement units, ensuring the control system operates within the reliable detection range of the attitude determination systems \cite{forbes2015fundamentals}.
\end{itemize}

\begin{figure}[H]  
    \centering
    \includegraphics[width=0.9\textwidth]{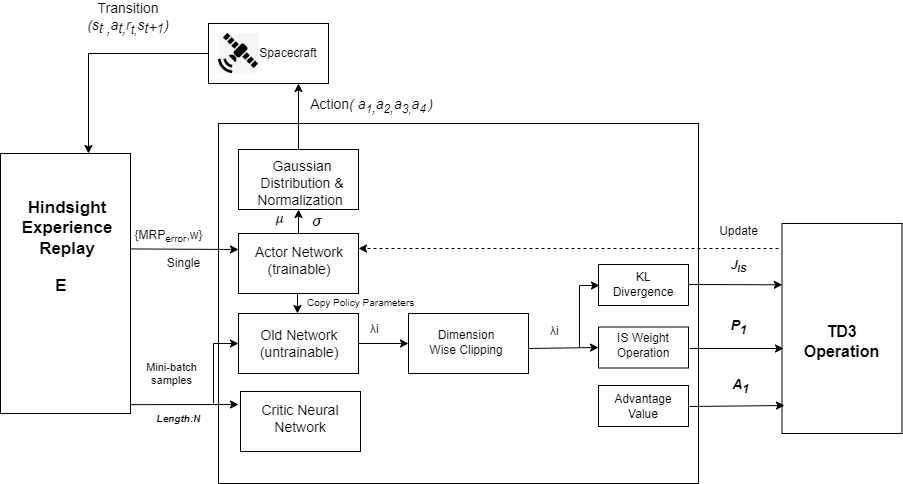}  
    \caption{Proposed TD3-HD  structure.}
    \label{fig:Proposed TD3}
\end{figure}

The overall reward function is:
\[
\text{reward} = r_1 + r_2 + r_3\tag{11}
\]
This function ensures that the spacecraft is rewarded for reducing attitude error, penalized for unsafe angular velocities, and incentivized to maintain accurate control.\\ \\
The TD3 algorithm, known for its stable learning and capability of handling continuous control problems \cite{al2024genetically}, is enhanced with Dimension-Wise Clipping (DWC), which independently limits action updates for each reaction wheel (RW). This mechanism helps mitigate large and unstable updates in the presence of faults \cite{han2019dimension}. Additionally, Hindsight Experience Replay (HER) addresses the issue of sparse rewards by reinterpreting failed actions as successes with modified goals \cite{xiao2023multimodal}, enabling the agent to learn effectively even in scenarios involving unresponsive RWs. The pseudo-code for the proposed method is outlined in Algorithm 1, and the network structure of the TD3-HD with DWC system is illustrated in Figure \ref{fig:Proposed TD3}. The system integrates two primary neural networks: an actor network and a critic network. The actor network, consisting of four sub-networks parameterized by $\lambda_i$ (for $i = 1, 2, 3, 4$), generates torque values for each RW by outputting Gaussian parameters $\mu_i$ and $\sigma_i$, which are sampled to produce normalized torque actions $a_i$ (for $i = 1, 2, 3, 4$). These actions are applied to the satellite's RWs to adjust its attitude. During training, the satellite's state, represented as $[\text{MRP}_{\text{error}}, \omega]$ (where $\text{MRP}_{\text{error}}$ denotes the Modified Rodrigues Parameters for attitude error and $\omega$ represents the angular velocity vector), is stored in the HER buffer $E$, which modifies failed trajectories by replacing goals with future goals. The TD3-HD method uses batch training, where the actor network updates its policy based on sampled transitions from $E$ and then copies its parameters to an "old" network to ensure stability. DWC is applied to the output of the old network, limiting large updates for each RW independently. Additionally, Importance Sampling (IS) weights $\rho_t$ are computed to correct for mismatches between the behavior policy and target policy, feeding into the TD3 operation. The critic network evaluates the generated torque actions by calculating an advantage value $A_1$, which reflects the quality of the actions. This advantage value, along with the IS weights and policy loss $J_{\text{IS}}$, is passed to the TD3 operation to finalize updates to the actor network. The system ensures efficient and stable training, resulting in updated Gaussian parameters $\mu_i$ and $\sigma_i$ for torque actions $a_i$, which are applied to the satellite's RWs to maintain and adjust its attitude.

\begin{algorithm}[H]
\caption{TD3-HD with DWC}
\begin{algorithmic}[1]

\STATE \textbf{Input:} Maximum iterations $L$, epochs $K$, HER goal sampling strategy, DWC bounds.

\STATE \textbf{Initialization:}
    \STATE Initialize actor network weights $\lambda_i$ for $i = 1, 2, 3, 4$ (corresponding to each RW).
    \STATE Initialize Q-value critic networks for TD3.
    \STATE Initialize Importance Sampling (IS) weighting factors $\alpha_{IS} = 1$, learning rate $\zeta$, and dimension-wise clipping thresholds for DWC.
    \STATE Initialize replay buffer $E$ for storing state transitions $\{(\text{MRP}_{\text{error}}, \omega), a, r, (\text{MRP}_{\text{error}}', \omega')\}$.
    \STATE Load the satellite dynamics model, including RW faults or unresponsive scenarios.

\STATE \textbf{Main Loop: Interaction Phase}
    \FOR{iteration = 1 to $L$}
        \STATE Initialize the satellite’s states $[\text{MRP}_{\text{error}}, \omega]^T$, including RW faults.
        \STATE Load the desired target states (goal orientation and angular velocity).
        \STATE Store current actor network parameters $\lambda_i \leftarrow \lambda$ (to retain a stable policy reference).
        \STATE Apply the actor policy $\pi_{\lambda_i}$ to sample actions $a = [a_1, a_2, a_3, a_4]$.
        \STATE Store state transitions $(s_t, a_t, r_t, s_{t+1})$ in the replay buffer $E$, applying the HER strategy to modify episodes by sampling future goals as desired goals.
        \STATE Compute advantage estimates $\hat{A}_t$ using transitions from the HER buffer $E$.
    \ENDFOR

\STATE \textbf{Training Loop}
    \FOR{epoch = 1 to $K$}
        \FOR{each gradient step}
            \STATE Sample a mini-batch of size $M$ from the replay buffer $E$.
            \STATE Apply dimension-wise clipping (DWC) to action updates $\lambda_i$ for each RW, limiting large and unstable updates.

            \STATE Update actor network weights:
            \[
            \lambda \leftarrow \lambda + \zeta \nabla_{\lambda} J_{\text{IS}}(\lambda)
            \]
            where $J_{\text{IS}}$ is the IS-weighted policy loss based on KL divergence.
        \ENDFOR
        
    \STATE Update the IS weighting factor $\alpha_{IS}$ adaptively to balance replay importance.
    \ENDFOR

\end{algorithmic}
\end{algorithm}

\section{Experimental Setup}

The Basilisk Astrodynamics Simulation Framework \cite{kenneally2020basilisk} provides a high-fidelity simulation environment for modeling the kinematics and dynamics of a small satellite in Low Earth Orbit (LEO). Hence, this simulator is used as simulation environment in our experiments The satellite’s orientation is represented by Modified Rodrigues Parameters (MRPs) and controlled using four Honeywell HR16 RWs arranged in a pyramid configuration. These RWs deliver torque adjustments to maintain the desired attitude of the satellite. To evaluate the fault tolerance of various control strategies, an RW failure is simulated by disabling one of the wheels at $3000^{th}$ second. Key telemetry data, including MRP, angular velocity, and RW torque, are recorded over a 8000-second period.

This study evaluates advanced DRL algorithms for CubeSat attitude control, focusing on the proposed TD3-HD algorithm. TD3-HD enhances control performance under actuator faults by integrating HER for improved learning in sparse reward environments. Additionally, Proximal Policy Optimization (PPO) \cite{tan2021mata}, known for its robustness in complex tasks and stable policy updates, is assessed alongside Advantage Actor-Critic (A2C) and the standard Twin Delayed Deep Deterministic Policy Gradient (TD3), which serves as a baseline for reliable CubeSat attitude control \cite{sarmiento2023sample}.

This comprehensive evaluation provides a comparative analysis of each algorithm's strengths and suitability for autonomous CubeSat attitude control.

\begin{figure}[H]
    \centering
    \includegraphics[width=0.7\textwidth]{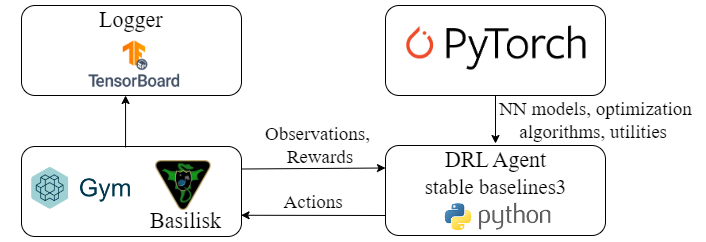}
    \caption{Simulation configuration used for satellite attitude control.}
    \label{fig:Setup}
\end{figure}

All DRL algorithms in this study are implemented using the Stable-Baselines3 library \cite{raffin2021stable}, an open-source toolkit built on PyTorch that facilitates efficient and reliable model development for DRL research. For training the neural networks, a custom simulation environment is created in Python, adhering to the OpenAI Gym standards \cite{1606.01540}. This standardization ensures a flexible interface for simulating interactions and enables seamless integration of the DRL algorithms. Figure \ref{fig:Setup} provides an overview of the simulation configuration.

The following table outlines the training parameters used for the TD3-HD algorithm. These parameters are used for optimizing the spacecraft attitude control system and ensuring effective learning under dynamic conditions.
TD3-HD exhibited stable and reliable performance using standard hyperparameter settings in (Table \ref{tab:td3her_parameters}) without the need for extensive tuning. The integration of HER and DWC contributes to this robustness by improving learning stability and mitigating the effects of sparse rewards. These characteristics make TD3-HD a practical and resilient choice for autonomous fault-tolerant satellite control.
\begin{table}[H]
    \centering
    \caption{Training parameters for the TD3-HD algorithm.}
    \begin{tabular}{@{}ll@{}}
        \toprule
        \textbf{Parameter} & \textbf{Value} \\
        \midrule
        Learning Rate (\( \zeta \)) & \( 3 \times 10^{-4} \) \\
        Replay Buffer Size & 1,000,000 \\
        Batch Size (\( M \)) & 128 \\
        Target Update Interval & 2 \\
        Clipping Thresholds (DWC) & 0.2 \\
        Hidden Units & 256 \\
        Importance Sampling (\( \alpha_{\text{IS}} \)) & 1 \\
        Trajectory Size (\( N \)) & 100 \\
        Actor Sub-networks (\( \lambda_i \)) & 4 \\
        \bottomrule
    \end{tabular}
    \label{tab:td3her_parameters}
\end{table}
\section{Experimental Results}
This section presents and analyzes the results of applying DRL to address the satellite attitude control problem. The performance of the control system was evaluated based on the pointing error between the satellite’s body-fixed and inertial frames, measuring the precision of orientation alignment.

Additionally, the performance of each approach was tested under a challenging actuator failure scenario (e.g., unresponsive RW). The findings demonstrate that, among the DRL algorithms, the proposed TD3-HD displayed a notably robust response, handling the critical failure scenario effectively and outperforming traditional methods in maintaining control accuracy and adaptability.

\subsection{Proportional-Derivative (PD) Controller: Analysis}

\begin{figure}[H]
    \centering
    \includegraphics[width=0.7\textwidth]{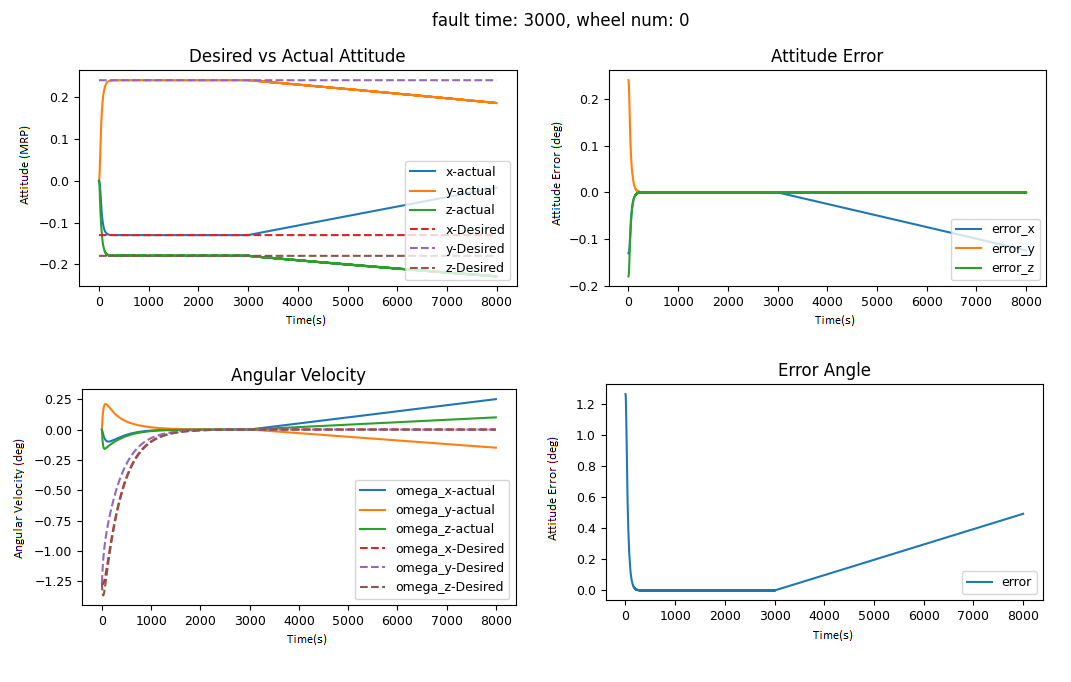}
    \caption{PD Controller Error Metrics for Attitude Control with RW Fault at 3000 seconds}
    \label{fig:PD_error}
\end{figure}

The results demonstrate the limitations of a PD controller in managing satellite attitude control following a RW failure. Initially, the PD controller effectively tracks the desired attitude, as shown by the "Desired vs. Actual Attitude" plot. However, following a fault in RW0 at $3000^{th}$ second, the PD controller fails to adapt autonomously, resulting in substantial deviations from the target attitude. The "Error History" and "Angle Error" plots (Figure \ref{fig:PD_error}) illustrate a sharp increase in attitude error after fault, with persistent oscillations of the angular velocity indicating that the controller’s parameters would require manual adjustment to restore stability. This lack of adaptability highlights a significant shortcoming of the PD controller under fault conditions.

\begin{figure}[H]
    \centering
    \includegraphics[width=0.7\textwidth]{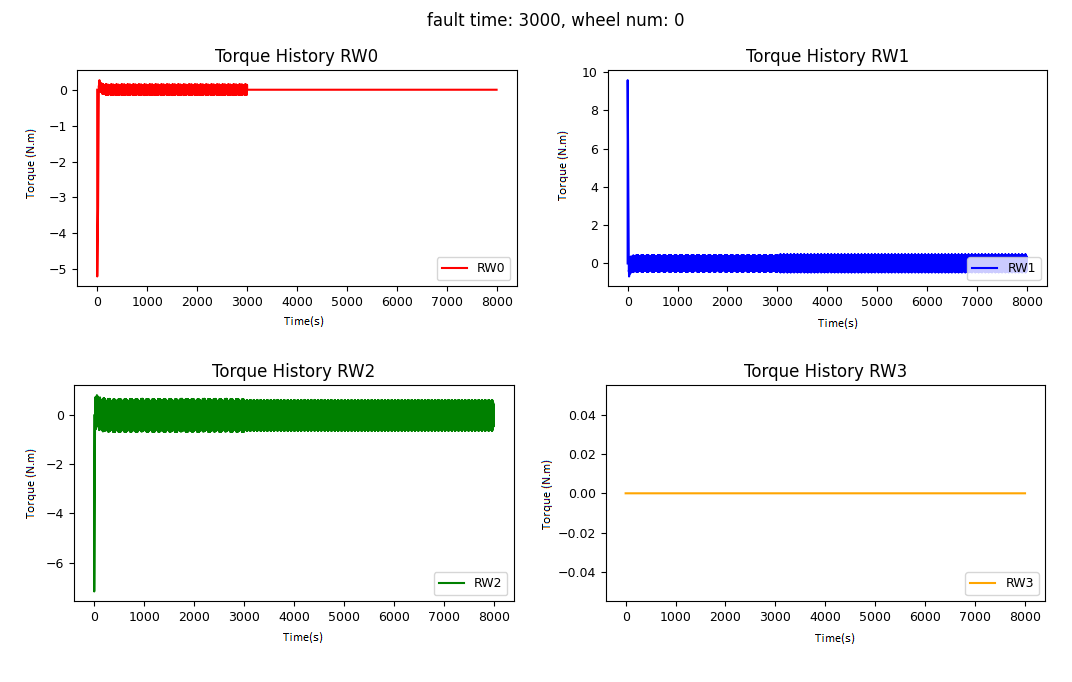}
    \caption{Torque History of RWs under PD Control with RW0 Fault}
    \label{fig:PD_error_torque}
\end{figure}

Additionally, the torque plots (Figure \ref{fig:PD_error_torque}) reveal that RW0 becomes unresponsive after the fault, leaving the remaining wheels to compensate without activating the backup wheel, which would require telecommand (ground intervention)  to retune the parameters. This limitation further destabilizes the system, highlighting the PD controller's inability to effectively manage fault recovery under such conditions.

These findings highlight the need for advanced, fault-tolerant control mechanisms, such as RL, which can autonomously adapt to actuator failures. Unlike traditional controllers, RL algorithms are capable of learning optimal control policies through experience, enabling real-time adjustments in response to faults and ensuring mission success without requiring manual intervention.

\subsection{A Comparative Analysis of RL-based approaches}

This section provides a comprehensive comparison of three DLR algorithms, A2C, PPO, and TD3, to address the unresponsive RW problem.

\begin{figure}[H]
    \centering
    \includegraphics[width=0.7\textwidth]{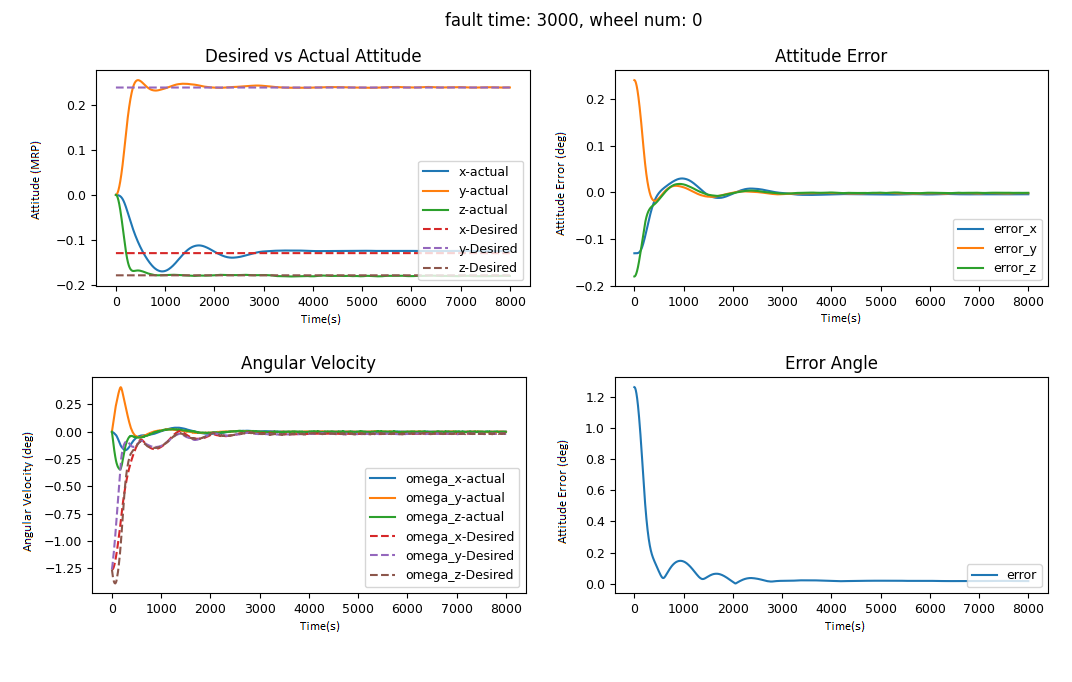}
    \caption{PPO Error Metrics for Attitude Control with Fault at 3000 seconds}
    \label{fig:PPO_error}
\end{figure}

\textbf{$\bullet$ PPO Performance Analysis:} The PPO algorithm shows effective adaptation following a fault in RW0 at $3000^{th}$ second. As illustrated in Figure \ref{fig:PPO_error}, the "Desired vs Actual Attitude" subplot, PPO successfully aligns the satellite's actual attitude with the desired orientation, minimizing deviations in a relatively short time. The "Attitude Error" subplot indicates that PPO maintains low error levels across all axes, even after the fault, demonstrating stable control. The "Angular Velocity" subplot reveals PPO’s capability to reduce the oscillations, achieving alignment of actual velocities with desired values.  However, PPO required more time to stabilize and experienced slight control variations due to exploration in rapidly changing conditions

\begin{figure}[H]
    \centering
    \includegraphics[width=0.7\textwidth]{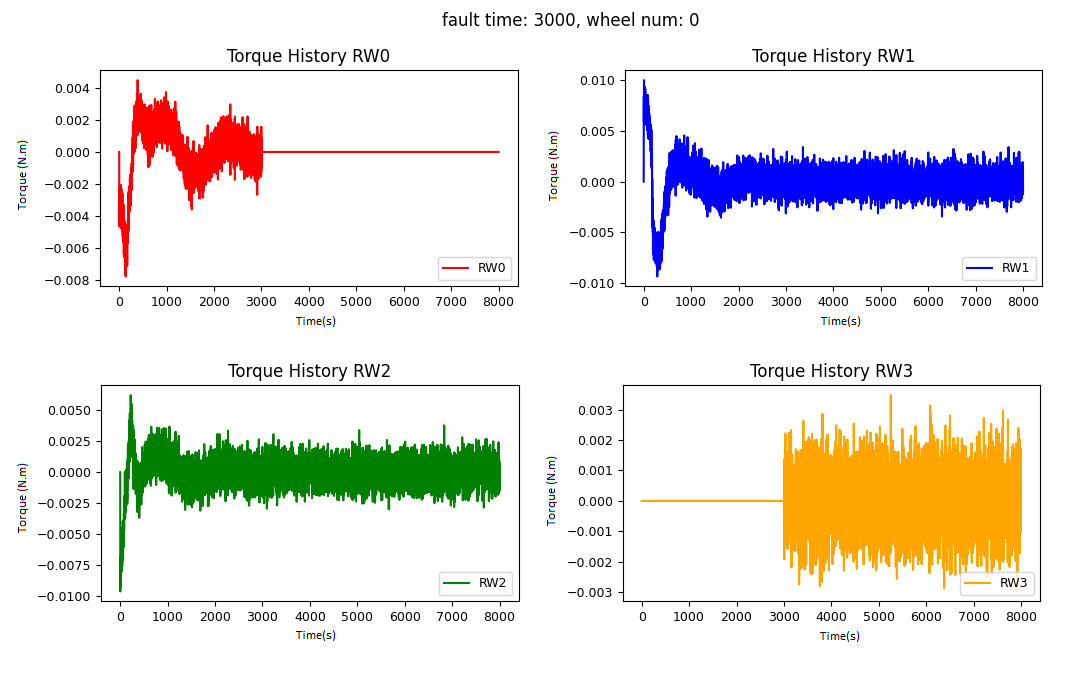}
    \caption{PPO Torque History for RW0, RW1, RW2, and RW3}
    \label{fig:PPO_torque}
\end{figure}

In terms of torque distribution, PPO dynamically redistributes torque among RW1, RW2, and RW3 after the RW0 fault, as shown in Figure \ref{fig:PPO_torque}. RW0’s torque drops to zero, while the other wheels exhibit increased torque fluctuations to compensate. This adaptive redistribution highlights the resilience of PPO in maintaining control and stability under fault conditions.

\begin{figure}[H]
    \centering
    \includegraphics[width=0.7\textwidth]{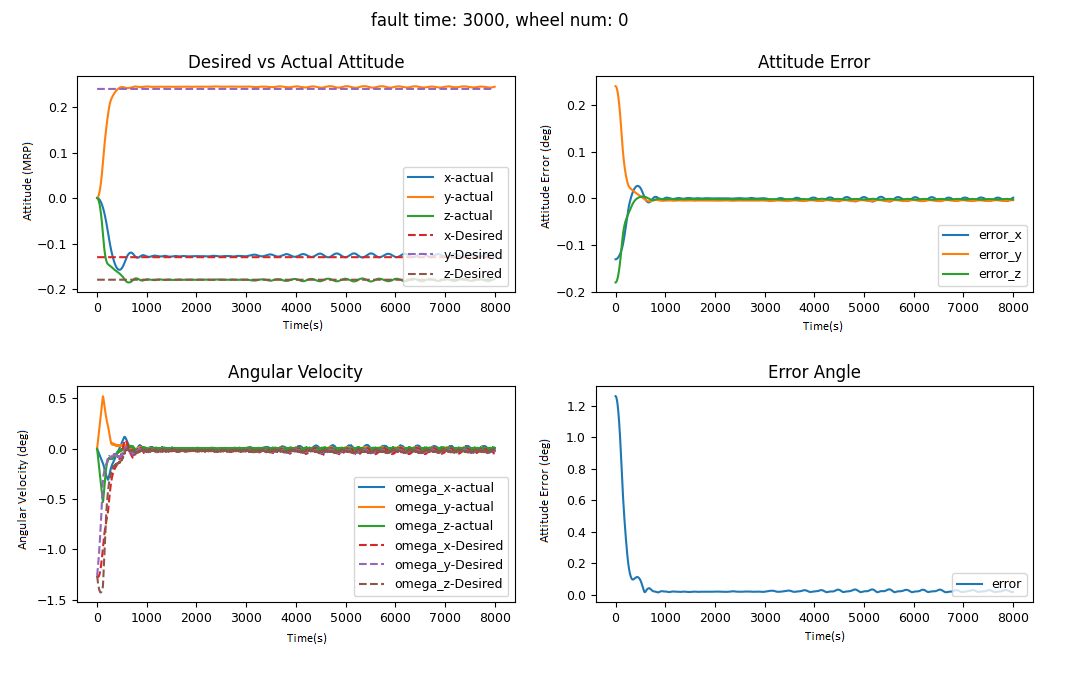}
    \caption{A2C Error Metrics for Attitude Control with Fault at 3000 seconds}
    \label{fig:A2C_error}
\end{figure}

\textbf{$\bullet$ A2C Performance Analysis:} The A2C algorithm also performs well following a fault in RW0 at $3000^{th}$ second. As shown in Figure \ref{fig:A2C_error}, in the "Desired vs Actual Attitude" subplot, A2C gradually minimizes deviations between the actual and desired attitudes, showing robust adaptability. The "Attitude Error" subplot indicates a rapid reduction in initial errors, particularly along the x-axis, with stable low error levels maintained post-fault. A2C effectively stabilizes angular velocities, as seen in the "Angular Velocity" subplot.  However, a limitation of PPO is that it requires more time to align with the desired attitude in highly dynamic environments, where it may struggle to respond quickly to changes in spacecraft orientation

\begin{figure}[H]
    \centering
    \includegraphics[width=0.7\textwidth]{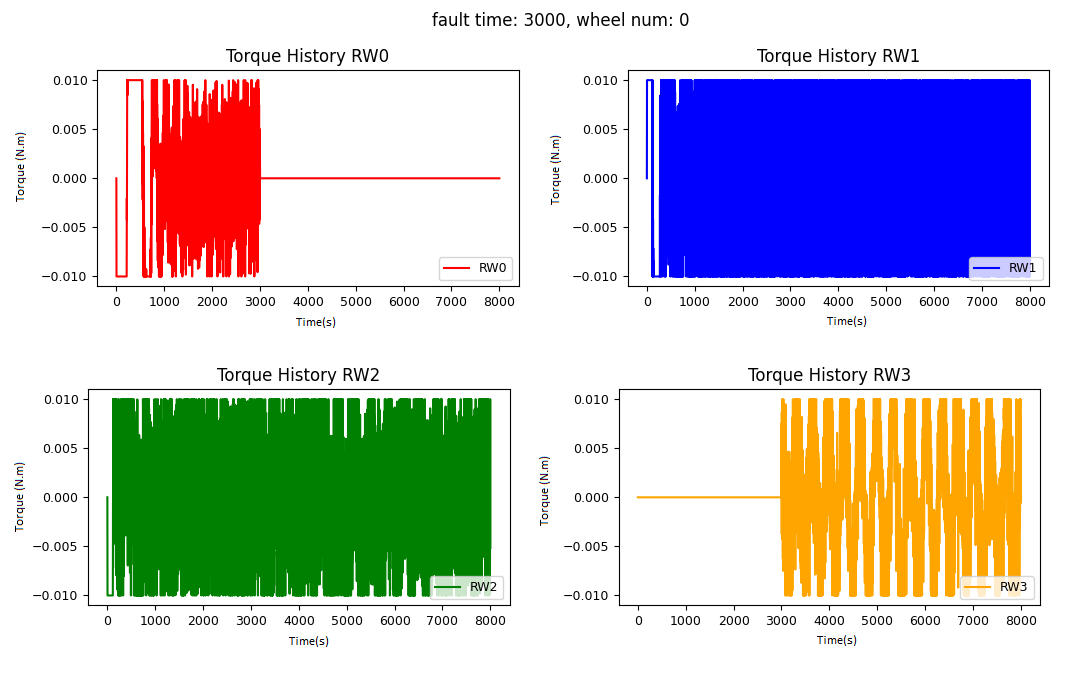}
    \caption{A2C Torque History for RW0, RW1, RW2, and RW3}
    \label{fig:A2C_torque}
\end{figure}

In the "Torque History" plots (Figure \ref{fig:A2C_torque}), A2C redistributes torque effectively among RW1, RW2, and RW3 to compensate for RW0’s inactivity.  

\begin{figure}[H]
    \centering
    \includegraphics[width=0.7\textwidth]{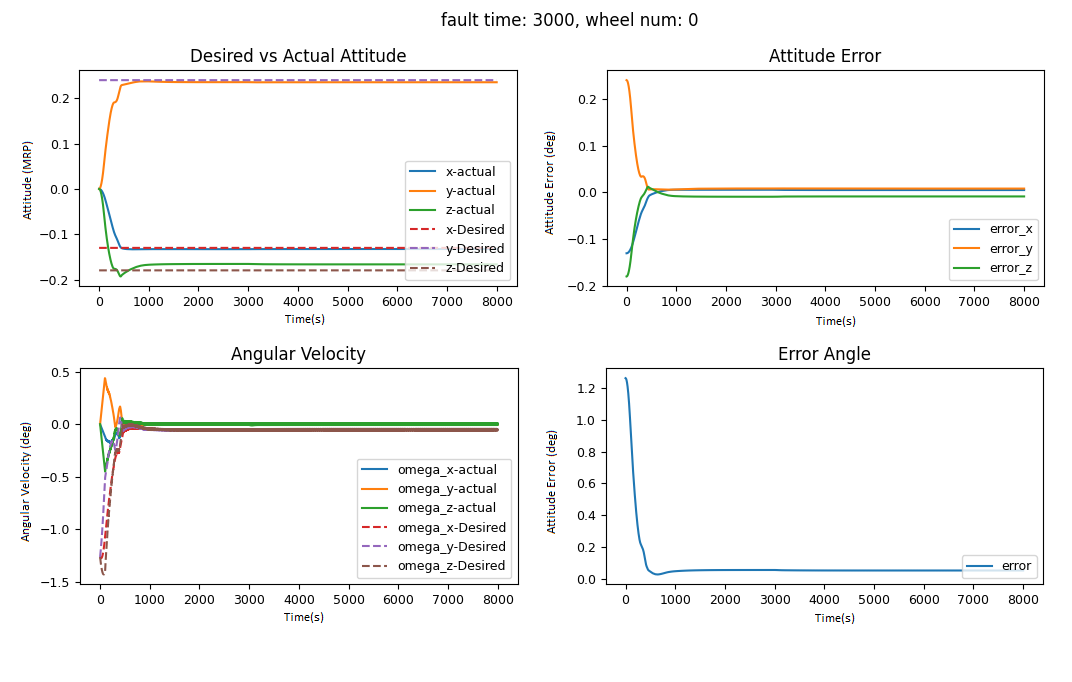}
    \caption{TD3 Error Metrics for Attitude Control with Fault at 3000 seconds}
    \label{fig:TD3_error}
\end{figure}

\textbf{ $\bullet$ TD3 Performance Analysis:} The TD3 algorithm demonstrates the strongest performance among the three, particularly in handling the RW0 fault introduced at $3000^{th}$ second. As illustrated in Figure \ref{fig:TD3_error}, the "Desired vs Actual Attitude" subplot reveals rapid alignment with the desired trajectory, with minimal deviations. In the "Attitude Error" subplot, TD3 achieves the quickest reduction in errors across all axes, maintaining near-zero error levels post-fault. The "Angular Velocity" subplot shows stable control with minimal oscillations, reflecting TD3’s high precision in managing angular velocities.

\begin{figure}[H]
    \centering
    \includegraphics[width=0.7\textwidth]{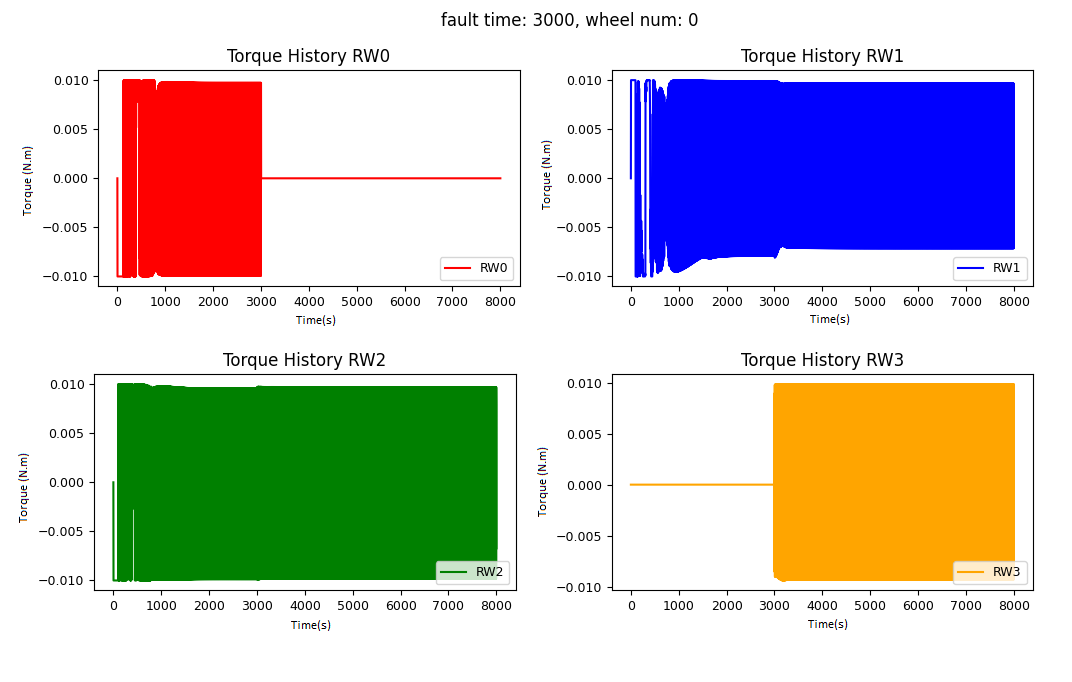}
    \caption{TD3 Torque History for RW0, RW1, RW2, and RW3}
    \label{fig:TD3_torque}
\end{figure}

TD3 achieves smooth torque redistribution in RW1, RW2 and RW3 after RW0 fault, as shown in Figure \ref{fig:TD3_torque}. This adaptive torque redistribution showcases the robustness of TD3 in ensuring control and stability during fault conditions.
However, the gradual convergence observed, particularly in the "Error Angle" subplot, reflects the challenges posed by sparse rewards, as the algorithm requires more iterations to achieve stable control.

\textbf{$\bullet$ Comparative Summary:} All three algorithms, PPO, A2C, and TD3, demonstrate effective fault tolerance in satellite attitude control, but each has limitations. PPO provides moderate adaptability, but can introduce oscillatory behaviors in persistent faults. A2C offers resilience and flexible torque redistribution, but has slower convergence and higher fluctuations. TD3 excels in precision and stability in both attitude control and torque distribution, though it requires higher computational resources and may be sensitive to over-adjustment. Overall, TD3 presents the most robust control solution, suitable for fault-tolerant applications, but the gradual convergence observed is attributed to the challenges posed by sparse rewards, which require more time for the algorithm to achieve stable control. These comparisons underscore the need for an advanced control mechanism that can overcome the issues mentioned above. The effectiveness of our proposed algorithm, TD3-HD, in addressing these challenges is evaluated and discussed in the following section.

\subsection{Performance of the Proposed TD3-HD}

\begin{figure}[H]
    \centering
    \includegraphics[width=0.7\textwidth]{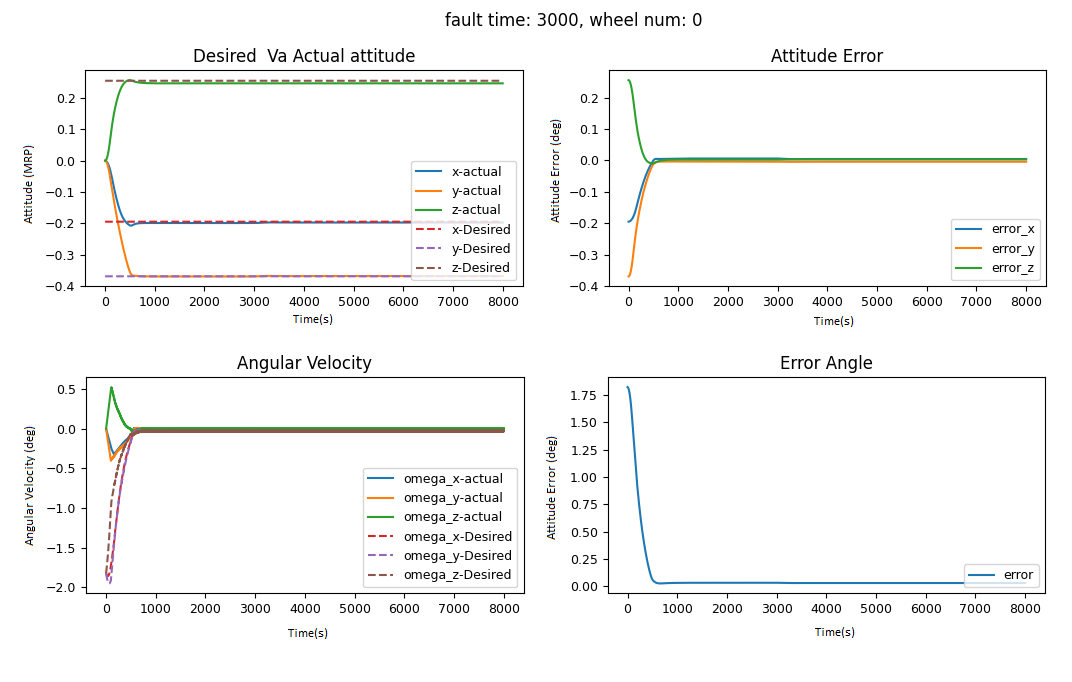}
    \caption{TD3-HD Error Metrics for Attitude Control with Fault at 3000 seconds}
    \label{fig:TD3_HER_error}
\end{figure}

\begin{figure}[H]
    \centering
    \includegraphics[width=0.7\textwidth]{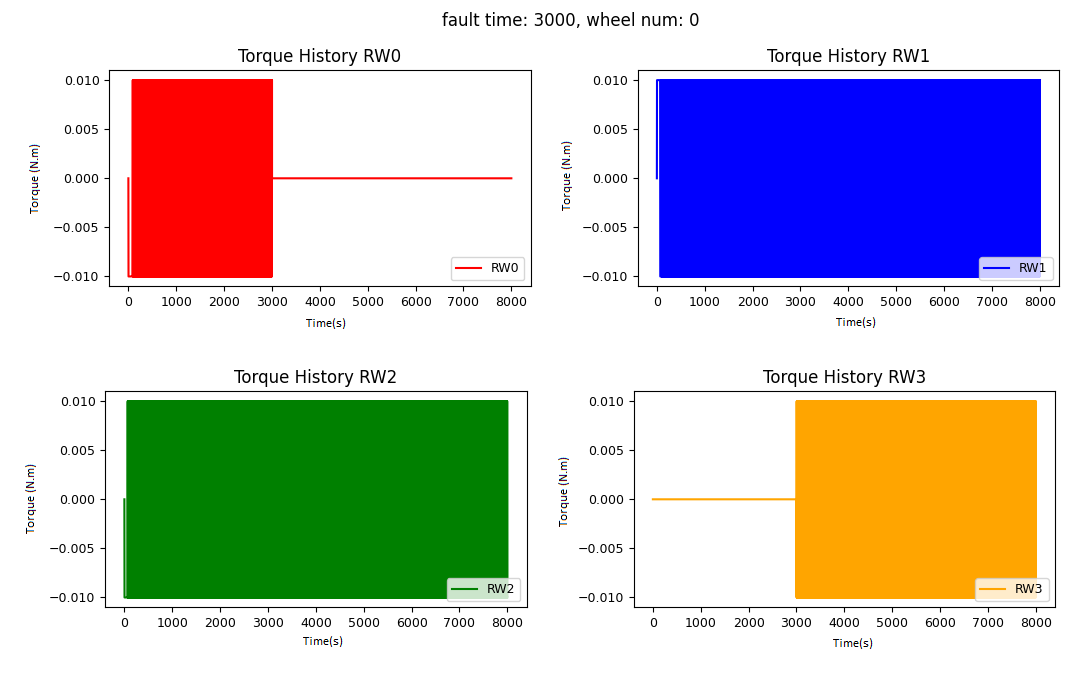}
    \caption{TD3-HD Torque History for RW0, RW1, RW2, and RW3}
    \label{fig:TD3_HER_torque}
\end{figure}

Figures \ref{fig:TD3_HER_error} and \ref{fig:TD3_HER_torque} display the performance of the proposed TD3-HD algorithm, incorporating HER and DWC. This enhanced TD3 variant addresses the limitations of standard TD3 and demonstrates superior performance in satellite attitude control. HER enables TD3-HD to learn effectively from sparse rewards, while DWC refines control precision by limiting torque adjustments dimensionally. These enhancements significantly boost resilience and adaptability under challenging conditions, such as the RW fault introduced at $3000^{th}$ second (RW0).

In the "Desired vs Actual Attitude" subplot, TD3-HD rapidly aligns the actual attitude with the target orientation, maintaining stability and precision throughout the simulation. The "Attitude Error" subplot reveals that TD3-HD minimizes deviations along all axes, achieving consistently low error levels post-fault, reflecting enhanced control accuracy. The "Angular Velocity" subplot demonstrates that TD3-HD effectively damps oscillations, achieving near-immediate alignment of actual and desired angular velocities. The "Error Angle" subplot further emphasizes the control precision of TD3-HD, with the attitude error angle quickly converging and remaining stable, underscoring the robustness of the algorithm in fault scenarios (Figure \ref{fig:TD3_HER_error}).

The torque history plots (Figure \ref{fig:TD3_HER_torque}) highlight TD3-HD’s efficient torque management. In "Torque History RW0," torque drops to zero following the fault, while the remaining wheels (RW1, RW2, and RW3) adjust smoothly and dynamically. DWC ensures dimensionally restricted torque outputs, preventing overcompensation and enabling balanced control. This approach providing smoother control than PPO, A2C, and standard TD3, as reflected in the consistent torque responses of RW1, RW2, and RW3.

Overall, the proposed TD3-HD algorithm, which integrates HER and DWC, offers exceptional adaptability and precision, effectively overcoming the limitations of existing DRL algorithms. TD3-HD establishes itself as a robust, efficient solution for satellite attitude control under fault conditions, achieving optimal torque distribution and attitude stability, surpassing the performance of PPO, A2C, and standard TD3.

\section{Findings}
The key findings of the research are summarized as follows:

\begin{itemize}
    \item \textbf{PD Controller Limitations:} The PD controller, though effective in maintaining the attitude of the satellite under nominal conditions, fails to adapt autonomously following a RW fault. After the failure of RW0 at $3000^{th}$ second, the PD controller exhibits substantial deviations from the target attitude and prolonged instability in angular velocity, necessitating manual tuning for stability restoration. This result highlights the PD controller's limitations in scenarios requiring fault-tolerant capabilities.

    \item \textbf{Comparative Analysis of PPO, A2C, and TD3:} The PPO and A2C algorithms provide satisfactory fault tolerance but exhibit certain limitations in control smoothness and convergence speed. PPO demonstrates moderate adaptability, but introduces oscillations in torque distribution under fault conditions. A2C offers flexible torque management, but shows slower convergence. TD3 excels in precision and stability, achieving rapid alignment with target orientations, although it requires higher computational resources and can be sensitive to over-adjustments in dynamic environments. Overall, TD3 proves to be the most robust of these three algorithms, though it still falls short of the proposed TD3-HD in terms of fault tolerance and efficiency.

    \item \textbf{The Proposed TD3-HD’s Enhanced Fault-Tolerance:} The proposed TD3-HD algorithm demonstrates superior fault tolerance and adaptability for satellite attitude control compared to other DRL algorithms such as PPO, A2C and TD3. Enhanced by HER, TD3-HD efficiently learns from sparse rewards, while DWC optimizes torque adjustments, minimizing fluctuations. This results in stable attitude alignment and low error levels even under RW fault conditions, effectively redistributing the torque without excessive oscillations. Overall, TD3-HD outperforms traditional and other DRL methods, making it the most effective autonomous solution for fault-tolerant control in satellites.
    \item \textbf{Implications for Autonomous Satellite Control:} This study underscores the potential of DRL, particularly the proposed TD3-HD, in enhancing satellite autonomy by adapting to actuator failures without manual intervention. This capability is essential for mission continuity, where reliability and adaptability are crucial for mission success in the face of unanticipated faults. The proposed TD3-HD’s ability to maintain control precision in challenging scenarios demonstrates its applicability for advanced space missions, require resilient and adaptive control solutions.
\end{itemize}

\section{Conclusion and Future Work}
This study introduced a DRL approach, called TD3-HD with DWC, to enhance satellite attitude control and address unresponsive RW faults in autonomous, fault-tolerant systems. TD3-HD’s combination of HER for improved learning from sparse rewards and DWC for stable individual torque adjustment, makes it well-suited for dynamic space environments. Simulations in the Basilisk environment showed that TD3-HD achieved lower attitude errors, better angular velocity control, and greater stability in RW fault scenarios compared to traditional PD control and other DRL methods (standard TD3, PPO, and A2C). This highlights the potential of TD3-HD as a robust, fault-tolerant, on-board AI solution for satellites, effectively enabling autonomous fault management and improving satellite resilience as an advanced autonomous system.
Future work will extend TD3-HD to satellite constellations, focusing on coordinated, fault-tolerant control across multiple satellites. Developing distributed DRL frameworks will support autonomous attitude management and adaptability to shared faults, advancing scalable, resilient operations for multi-satellite environments.

\section*{\uppercase{Acknowledgements}}

This work has been supported by the SmartSat CRC, whose activities are funded by the Australian Government’s CRC Program. We also acknowledge the collaborative contributions of the research team (James Barr, Travis Bessell) from Saab Australia.

\clearpage
\bibliographystyle{elsarticle-num} 
\bibliography{References}

\end{document}